# How False Data Affects Machine Learning Models in Electrochemistry?

*Krittapong Deshsorn[a,b], Luckhana Lawtrakul[a], Pawin Iamprasertkun[a, b,]\**

[a]*School of Bio-Chemical Engineering and Technology, Sirindhorn International Institute of Technology, Thammasat University, Pathum Thani, Thailand 12120 (pawin@siit.tu.ac.th)*

[b]*Research Unit in Sustainable Electrochemical Intelligent, Thammasat University, Pathum Thani, Thailand 12120*

TOC: Graphical Abstract

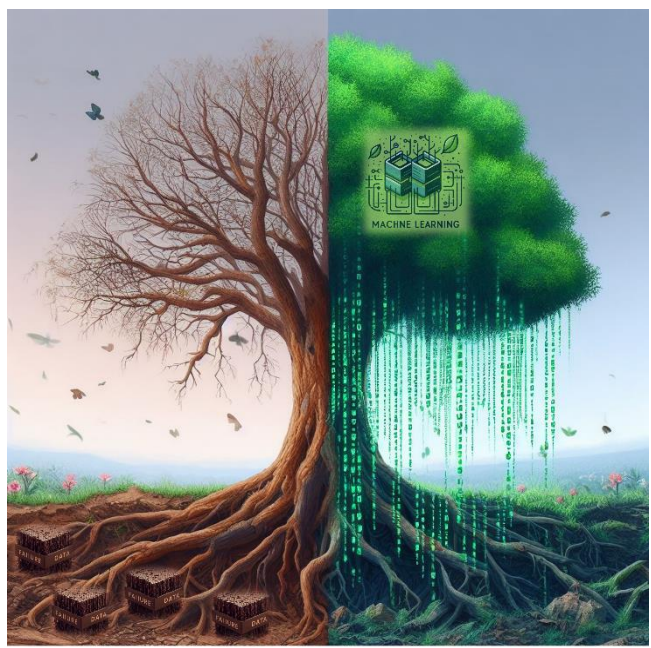

ABSTRACT

Noisy data is detrimental to the prediction of machine learning in chemistry. But some models are more tolerant to noise than others. The selection of machine learning models in electrochemistry is based on only the data distribution without concerning the noise of the data. This study aims to provide a discussion of the introduction of failure input data in electrochemistry, which demonstrated using heteroatom doped graphene supercapacitor data. The electrochemical data were tested with 12 standalone models including XGB, LGBM, RF, GB, ADA, NN, ELAS, LASS, RIDGE, SVM, KNN, DT, and our "stacking" model. By gradually adding the noise/false data into the pool, the models were then trained on both noisy and ground truth data to obtain various error metrics (MAE, MSE, RSME, MAPE, and $R^2$). The linear regression was then fitted on the increasing/decreasing errors to obtain the slope and intercept to discover estimated base accuracy (intercept) and noise sensitivity (slope). Hence, this study utilized contour plots, SHAP, and PDP to explain how the error affects the electrochemical feature including prediction and analysis. It is found that linear models handle the false data well with an average MAE slope of 1.513 F $g^{-1}$, but it suffers from prediction accuracy due to having an average MAE intercept of 60.20 F $g^{-1}$. This is due to improper model selection for this type of data (average $R^2$ intercept of 0.25). The "Tree-based" models fail in terms of noise handling (average MAE slope is 58.335 F $g^{-1}$), but it can provide higher prediction accuracy (average MAE intercept of 30.03 F $g^{-1}$) than that of linear models. Tree-based models also fit well to the data (average $R^2$ intercept of 0.9516). This suggests that the linear based model can be well described the relationship between capacitance and surface area. While the "tree based" model can be used for handling the other electrochemical features e.g. amount of heteroatom doped, current density, and so on. Miscellaneous models such as SVM, KNN, and NN, are moderately robust to noise (average MAE slope of 25.956 F $g^{-1}$) and provide moderate accuracy (average MAE intercept of 41.306 F $g^{-1}$). The models also fit moderately well to the data (average $R^2$ intercept of 0.546). To address the controversy between prediction accuracy and error handling, the "stacking model" was constructed, which not only shows high accuracy (MAE intercept of 24.29 F g$^{-1}$), but it also exhibits good noise handling (MAE slope of 41.38 F g$^{-1}$ and $R^2$ intercept of 0.86), making stacking models a relatively low risk and viable choice for electrochemist. This study presents that untuned NN is not suitable for electrochemical data, and improper tuning results in a model that is susceptible to noise, which directly affects the misleading in the electrochemical discussion. Thus, "STACK" models should provide better benefits in that even with untuned base models, it can achieve an accurate and noise tolerance. Overall, this work provides insight into machine learning model selection for electrochemical data, which should aid the understanding of data science in chemistry and energy storage context.

INTRODUCTION

The "Fourth Industrial Revolution" signifies a profound shift in how we live, work, and conduct research. This new phase of human advancement, driven by remarkable technological progress akin to the first three industrial revolutions, marks a significant transition. These technological transitions make data become more significant ("Data is new gold") in all areas. As a chemist, we could observe a significant rapid growth in the interdisciplinary between chemistry and data science[1]. Many machine learning works were studied during the past decade using a variety of model e.g., neural network[1, 2], tree-based[2, 3], kernel[4], linear[5] for solving chemistry tasks[2-4]. It requires tons of data for an accurate prediction (forecast); therefore, the data must be gathered during these processes. Some literature use either text mining or large language models (LLMs e.g. GPT-3.5 and GPT-4.0)[6] alternatively to manual data collection[7] for obtaining enough data points. Even though those methods are very useful tools as an "AI-powered" chemistry assistant, the validity of the data, which strongly influence the results must be considered. In fact, error (noise) in the data has always been a hurdle in the development of machine learning models. Error can be caused by imperfect data collections processes e.g., wrong measurement, uncontrolled laboratory conditions, wrong data logging, and false (also referred to fake and manipulated) data[8]. This imperfection in data collection is unavoidable, but steps can be taken to minimize or prevent errors[9-11]. When this data is used to train machine learning models, irrelevant and unnecessary information causes the prediction of the model to deviate from the ground truth. It is shown that noisy data leads to unreliable and inaccurate predictions[12,13]. Therefore, the utilization of models, which are robust to noise and provide high accuracy is a must in all fields. Especially in the energy storage context, where data is scarce, collection from various sources is common[14-16]. This can introduce noise into the data as various experiments inherit different conditions (i.e., temperature, time, batch). This means that not only the highest accuracy model is needed, but the model must inherently exhibit exceptional noise handling (data failure) or utilize techniques to reduce those errors.

In this case, an error can provide certain "uncertainty" in the outcome of a noisily trained model's prediction. The concept of "noise" linking with uncertainty has been studied in the concept of "aleatoric" and "epistemic" uncertainty[17]. The aleatoric uncertainty refers to uncertainty that is an inherent property of the collected data (e.g., noise, class overlaps). It is an irreducible error in the data collection process that persists even with additional data[18]. Epistemic uncertainty, unlike aleatoric uncertainty, is a reducible error caused by the general lack of knowledge of the best tuned model or insufficient data to train the model. This uncertainty can be reduced by optimizing model hyperparameters or finding more data[18]. Many machine-learning studies in chemistry have been focusing on reducing epistemic uncertainty for obtaining best prediction[19-21]. An example of studies that focus on tuning is a study from Zhang *et al.*, they predicted the compressive strength of alkali activated materials by focusing on the model optimization (reducing epistemic uncertainty). A heavy focus on hyperparameter tuning using Bayesian optimization algorithm (BOA) was applied, then feature importance was obtained[19]. Qi *et al.* collected their data from various sources, in which a model is tuned to obtain high accuracy to finally apply SHAP analysis and obtain feature importance[20].

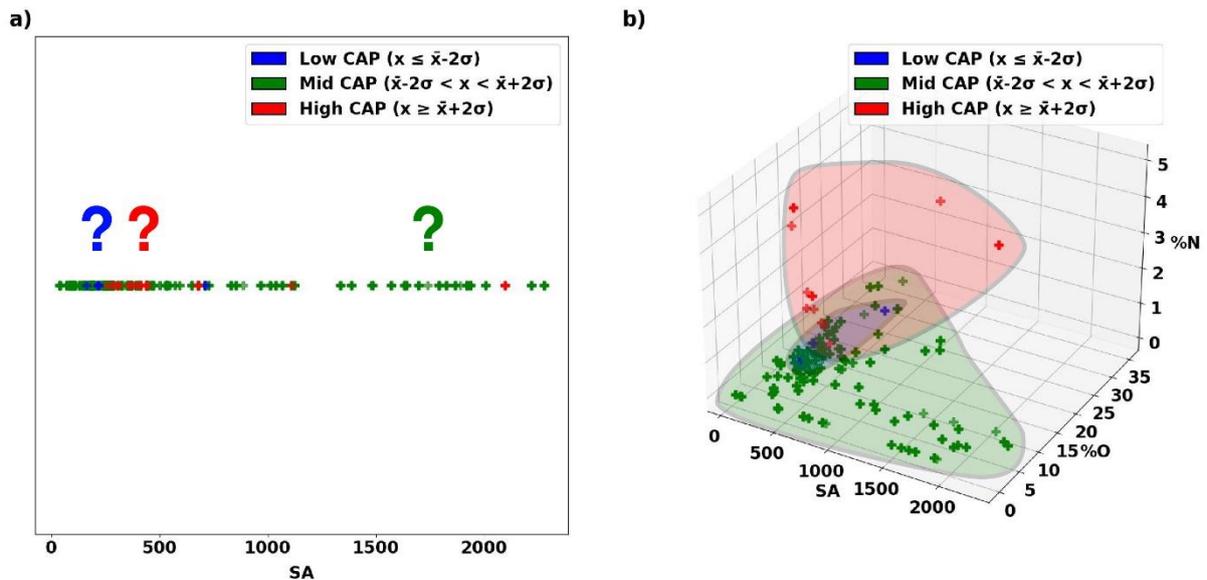

**Figure 1**: Examples of the data distribution for the demonstration of aleatoric vs. epistemic uncertainty (a) overlapping error/aleatoric uncertainty, and (b) non-overlap/epistemic uncertainty.

To visualize aleatoric/epistemic errors in electrochemistry aspect, **Figure 1** shows a classification task for supercapacitor data from our previous work[7, 22]. The blue, gray, and red data points represent low, medium, and high capacitance, respectively. When only one feature is used to separate capacitance groups (see **Figure 1a**), overlaps of data cause uncertainty in decision boundaries for the data separation. This causes difficulty in machine learning model predictions. This uncertainty cannot be reduced even with the addition of more data, as the noise and irrelevant patterns would remain. When three features are used in **Figure 1b,** the uncertainty due to class overlap is then reduced. As seen from **Figure 1b**, the red class is now easier to classify than the previous situation (refer to **Figure 1a**). This reduces aleatoric uncertainty, but in turn increases epistemic uncertainty. Meaning that the uncertainty comes from unknown model parameters and proper models. Thus, a well-defined hyperparameter and appropriate machine learning model is then required to solve this problem[14].

However, the focus of only the parameters tuning i.e., reducing epistemic uncertainty, may not address the reduction of aleatoric uncertainty and cause overfitting from over-tuning by adding too much complexity[23]. Therefore, a broader study into aleatoric uncertainty handling can act as a "filter" to search for a suitable model; hence, epistemic uncertainty can be reduced by tuning parameters. To the best of our knowledge, two distinct ways can be used to minimize aleatoric uncertainty for machine learning, that is to reduce it from the "data" perspective and "model" perspective. From the "data" perspective, aleatoric uncertainty is minimized by utilizing techniques that lower the amount of noise in the data [24,25,26,27]. For example, Nicola Segeta *et al.*, introduced a way to reduce noise (i.e., data point in wrong domain) by training a local support vector machine to identify whether the data point should be removed or not[28]. This can remove mislabeled data, essentially reducing noise. Another work is noise reduction by limiting the influence of outliers by Miroslaw Kordos *et al*. This limits the outlier's impact on the model without rejecting the data point, essentially keeping important information that may otherwise be wasted, if removed[27].

From the "model" perspective, techniques can be utilized to increase a model's ability to handle noise, for example, methods such as bagging (e.g., RF[29]) and boosting (e.g., ADA[30]) can be utilized to handle noise. RF essentially handles noise by randomly boot strapping data[31] and building multiple trees trained on those boot straps. This can reduce error by randomly excluding noisy data from the training set and using multiple predictors. ADA handles noise by utilizing weak learners and assigning weights to training samples. The weights determine the probability that each data point will be included in the next weak learner's training set. The weight of the misclassified example will increase, causing the next weak learner to hopefully be trained and better identify the problematic data points. This can reduce noise by essentially handling noisy data points down the pipeline. Another interesting ensemble technique called the "stacking" technique, which combines many base models and meta models, is still poorly understood in its ability to handle noise compared to bagging and boosting. Therefore, this study aims to study how the "stacking" technique compares to other standalone models when trained on noisy supercapacitor data. A real-life scenario, where the capacitance sometimes is over-inflated due to technical errors and how it affects the prediction is also studied. Our studies concludes that the "stacking" model, is both a great generalizer, and has high prediction accuracy[22], but can fail in terms of feature importance analysis if improper base models are selected.

Combining these properties with a robustness to noisy data and inflated data can yield a generalized model that has less chance of failure for the general public and research sectors alike. Thus, this paper aims to demonstrate the aspect of error handling of "stacking" models and various commonly used models using energy storage data. Several standalone models were tested, which includes extreme gradient boost (XGB), light gradient boosting machine (LGBM), random forest (RF), gradient boost (GB), adaptive boosting (ADA), neural networks (NN), elastic-net (ELAS), lasso regression (LASS), ridge regression (RIDGE), support vector machine (SVM), k-nearest neighbors (KNN), and decision trees (DT), totaling 12 models. The stacking model was then constructed with a combination of the standalone models as the base layer and linear regression "meta" layer (secondary layer). The model's analysis is separated into four groups of models, which are: (1) Linear family such as ELAS, LASS, RIDGE, (2) Tree-based model including DT, RF, GB, ADA, XGB, LGBM, (3) Miscellaneous, which are

NN, SVM, and KNN, and (4) the as-constructed stacking model. To the best of our knowledge, we have found that the linear models are robust to noise, but achieve low prediction accuracy, tree-based models are not robust to noise, but achieve high prediction accuracy, NN is robust to noise but must be tuned properly to achieve high accuracy (major awareness when apply NN model), SVM is both moderately robust and accurate, and KNN is susceptible to noise but provides moderate accuracy. Finally, the STACK model is the winner in these games. The model is robust to noise and performs well among all the models. This study aims to reduce the risk of choosing and tuning various models and parameters, also helping the chemical community and general public in providing useful insight into models and techniques that are inherently robust to noise. This also allows confidence in beginners and experts alike in the machine learning space to use models for prediction and feature importance analysis.

METHODOLOGY

The methodology in this study is divided into 3 parts. The data and data analysis, perturbation of data, and model development, respectively. The data in this study is obtained from our previous study[22]. A summary is that the data is collected from various literatures and accumulated into a CSV. The data must be related to graphene-doped supercapacitors. The methods used to collect and clean the data can also be found in the previous study, which includes imputation and scaling of the data. In total, we have used 620 datapoints with seven input features and one target result, which is capacitance. The dataset is available as a CSV attached to the literature. For elementary data analysis, a brief overview of the scattered data in Figure S1, which is found in the Supplementary Information, suggests that a simple model like multiple linear regression may not accurately predict the relationship between the capacitance and features as the data is extremely scattered and contains a lot of noise (i.e., same y-value in the same x-value). This is due to the inherent pattern in the data as many features can be the same but have varying current densities and electrolyte concentration. This points to the issue of multicollinearity in the data as seen in the heatmap of Figure S2. Multicollinearity is a problem in machine learning in that it can cause varying and biased results in the model[32]. This is due to changes in one feature affecting another, which affects the explanation and relationship of the feature on the target variable[33]. As seen in Figure S2, some features are highly correlated, such as SA and %O (Pearson of -0.35), %N and %S (Pearson of 0.51), etc. Coincidentally, some models are robust to multicollinearity. These models include tree-based models[34] as they are based on binary recursive splits, meaning they handle data individually at a time instead of large clusters, causing multicollinearity to be less effective on the model. Another interesting model to handle multicollinearity is NN[35], in that deep architectures can learn extremely complex relationships, which can drown out the multicollinearity in layers.

To simulate aleatoric uncertainty, noise must be added to a dataset. For numerical data, noise is added by multiplying each data point in the dataset by a percentage in a range. For example, each data point in a dataset is randomly changed by a range of -X% to X% using a uniform distribution. This essentially simulates noise in real world numerical data (e.g., wrong measurement, incorrect data keying, falsifying data). An example can be seen in Figure S3, where noise is randomly added into numerical data. In this study, 10 noise levels are tested which are from 0% to 100% with a 10% interval. The perturbation is done on both features and target, which is reported. For only target perturbation or feature perturbation results, the results can be found in Supplementary Information (zipped). The overall workflow in this study can be found in Figure 2. The models used consists of 12 models and a stacking model, which includes XGB, LGBM, RF, GB, ADA, NN, ELAS, LASS, RIDGE, SVM, KNN, DT, and the stacking model. The stacking model consists of all 12 models as the base model. The meta model (i.e., the second layer) for the stacking model is a simple linear regression model. All models and the NumPy seed were fixed at a random state of 0 (if applicable) for reproducibility. The models are not tuned in anyway, as focus on general use and aleatoric uncertainty is the goal. The default parameters can be found in their respective libraries and summarized in Table S1 where the data distribution and Peason plot are shown in Figure S1 and Figure S2. For the explanation and workflow of the stacking process, refer to Figure S3 and Figure S4.

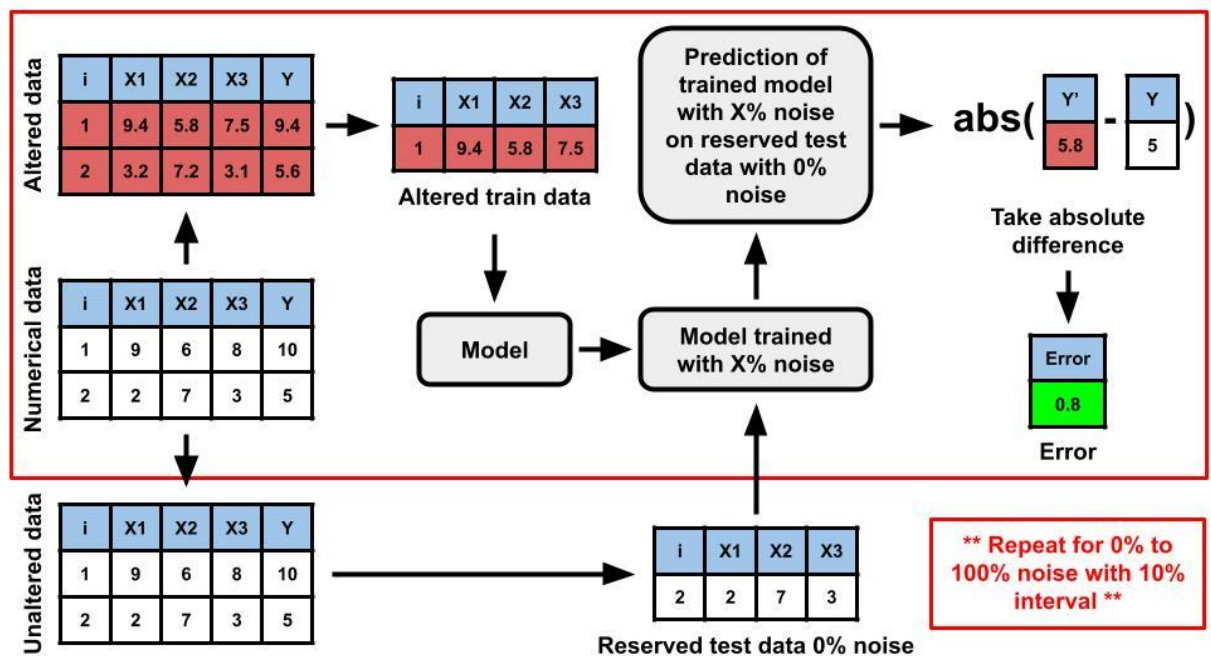

Figure 2: The calculation workflow used in this study for the mean absolute error (MAE) metric.

The workflow of each model is as follows:

1) The dataset is copied into 2 datasets. The 2 datasets would become the unaltered dataset (ground truth) and the altered dataset (perturbed).

2) The perturbed dataset is preprocessed (e.g., changed by random value, imputation, cleaning) and split into a train and unused data set. The unused data set should be the test set, but the test set will be the unaltered data that is split using the same randomization index.

3) The model is trained on the perturbed train dataset to simulate a model that is trained with noisy data.

4) The test set, which is the unaltered dataset split using the same randomization index, is fed to the model to predict the capacitance that would be obtained from a model trained.

5) The prediction is compared to the true values to get the error of the model.

6) The data is compiled into a CSV.

7) The process is repeated for all randomization values 0% to 100% with a 10% interval. Each interval is also repeated for 10 randomization seeds, ranging from 0-9, which allows reproducibility and reliability.

To establish how well each model handles noise (i.e., how the false data affect the prediction result), metrics for evaluation must be obtained, in which the metrics used were mean absolute error (MAE), mean squared error (MSE), root mean squared error (RSME), coefficient of determination ($R^2$), and mean absolute percentage error (MAPE). The mathematical equations for those criteria can be found in Eqn. 1 to Eqn. 5, respectively.

$$MAE = \frac{\sum_i |y_i - \hat{y}_i|}{n} \qquad \text{(Eqn. 1)}$$

$$MSE = \frac{\sum_i (y_i - \hat{y}_i)^2}{n} \qquad \text{(Eqn. 2)}$$

$$RMSE = \sqrt{\frac{\sum_i (y_i - \hat{y}_i)^2}{n}} \qquad \text{(Eqn. 3)}$$

$$R^2 = 1 - \frac{RSS}{TSS} = 1 - \frac{\sum_i (y_i - \hat{y}_i)^2}{\sum_i (y_i - \bar{y})^2} \qquad \text{(Eqn. 4)}$$

$$MAPE = \frac{\sum_i \frac{|y_i - \hat{y}_i|}{y_i}}{n} \qquad \text{(Eqn. 5)}$$

where RSS is sum of squared residuals, TSS is total sum of squares, $y_i$ is the real value, $\hat{y}_i$ is the predicted value, and $\bar{y}$ is the mean of the real values, and n is the number of observed points.

RESULT & DISCUSSION

The two main metrics e.g., MAE and $R^2$ are presented in the main text, the other criterions were reported in Supplementary Information. Remarks, for XGB at 100% noise, a data point has an extremely high MAE (540.5748 F g$^{-1}$) and low $R^2$ (-51.1984), therefore, the range of the plot has been adjusted so as not to skew the plot. For additional metrics, namely mean squared error (MSE), root squared mean error (RSME), mean absolute percentage error (MAPE), and unadjusted plot for MAE and $R^2$, refer to Figure S5 to Figure S9, respectively. Overall, the noise percentage is plotted in the x-axis and MAE of the prediction compared to the ground truth is plotted in the y-axis. Figure 3 shows the performance of all models. Each noise percent in the x-axis has 10 dots in the y-axis, which represents the 10 seeded splits done. As observed, when the percentage of noise is increased, the prediction error deviates more from the ground truth. By fitting a simple linear regression onto the data, the observation of deviation with respect to error can be obtained. The higher slope of the linear regression line indicates that the model is less robust to noise (low noise handling/false data strongly affect the prediction performance). The intercept of the line indicates the estimated lowest error present when there is no false data or noise in the dataset. Higher intercept indicates less accurate predictions from the model. For example, comparison of RIDGE and DT suggests that RIDGE (slope equals 4.11 F g$^{-1}$) can handle error in the data better than that of DT (slope equals 102.11 F g$^{-1}$) due to less slope. However, the DT (intercept equals 32.29 F g$^{-1}$) provides higher prediction accuracy than RIDGE (intercept equals 53.85 F g$^{-1}$). Now that the technique has been established, comparison of all models tested in this study can be done. Overall, this technique can serve as a "filter" for which models handle aleatoric error well for the specific dataset, and epistemic error reduction (i.e., tuning the model) can be done afterwards to ensure high accuracy results. When using MAE as the metric, the linear model provides high resistance to failure in the data (slope of ELAS, LASS, and RIDGE is 0.08, 0.35, 4.11 F g$^{-1}$ respectively) but they show low prediction accuracy (intercept of ELAS, LASS, and RIDGE is 63.42, 63.34, 53.85 F g$^{-1}$, respectively). Obviously, these models are great for non-complex data and generalization, unlike electrochemistry data. The tree-based group is extremely sensitive to noise (slope of LGBM, XGB, GB, RF, and DT is 45.36, 87.71, 57.31, 40.68, 102.11 F g$^{-1}$, respectively) with a notable exception of ADA (slope is 16.84 F g$^{-1}$), which is more tolerant to noise. In contrast, they can provide high prediction accuracy when there is no noise in the data, therefore, the input data must be perfect. Those tree-based models (LGBM, XGB, GB, RF, and DT) exhibit intercepts of 26.05, 17.01, 29.49, 27.68, 32.29 F g$^{-1}$, respectively. Therefore, the tree-based models are excellent for users that aim for highly accurate predictions, however, they must be ensure that the data has no defect. In the miscellaneous category, SVM provides a good balance between noise tolerance (slope of 27.13 F g$^{-1}$) and prediction accuracy (intercept of 40.93 F g$^{-1}$), making it a good choice for electrochemical data. KNN is more sensitive to noise (slope of 41.06 F g$^{-1}$) but has good accuracy (intercept of 33.28 F g$^{-1}$) and does not need to be heavily tuned as there are few parameters. NN in this case provides tolerance to noise (slope of 9.68 F g$^{-1}$) but has low prediction accuracy (intercept of 49.71 F g$^{-1}$). It should be noted that the NN cannot be used straight away for most datasets but must be properly tuned and managed. Our "STACK" model with the advantage of all the models presents good noise handling (slope of 41.38 F g$^{-1}$) and high prediction accuracy (intercept of 24.29 F g$^{-1}$).

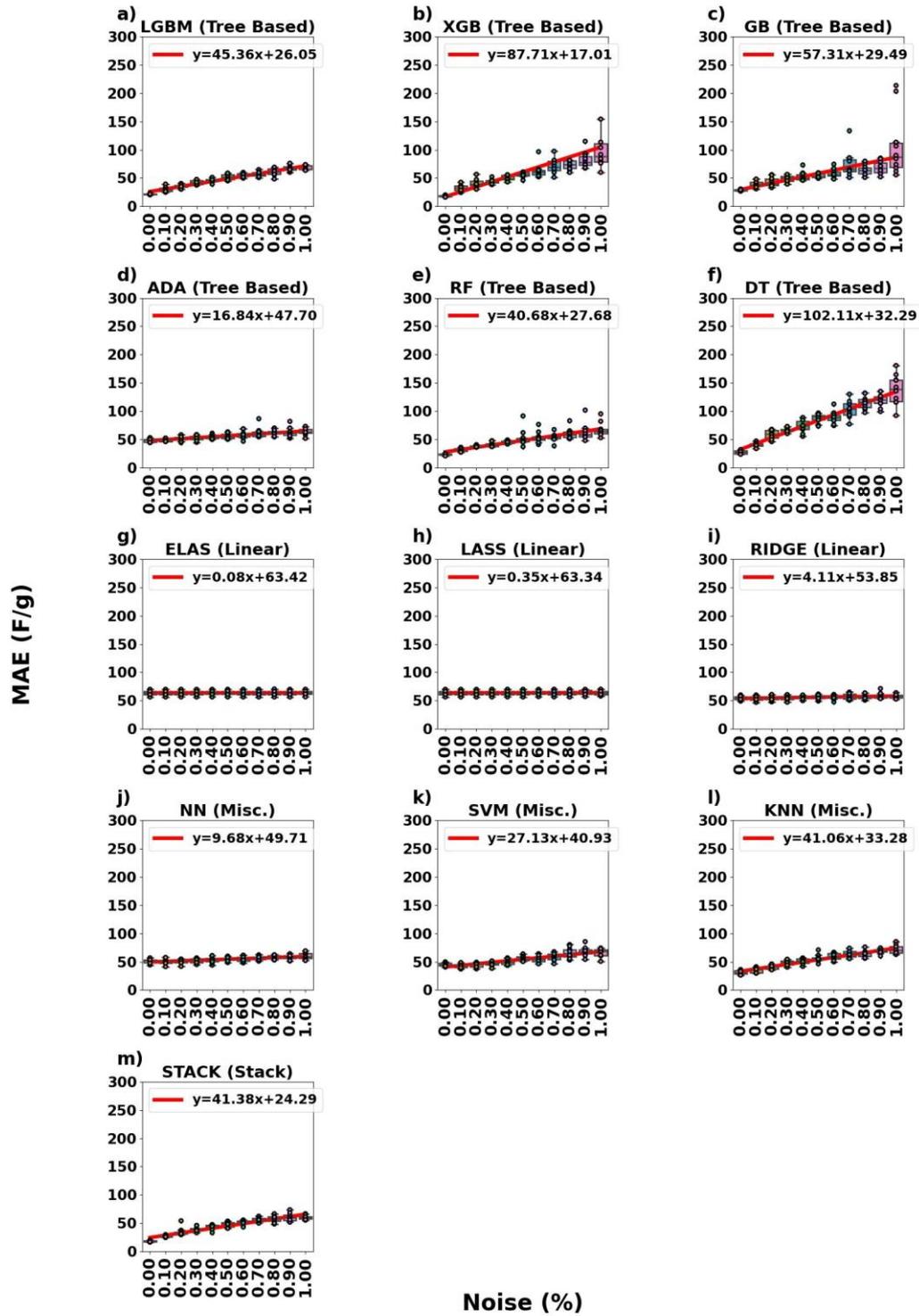

Figure 3: The increase in MAE of each model when features and target are perturbed for (a) LGBM, (b) XGB, (c) GB, (d) ADA, (e) RF, (f) DT, (g) ELAS, (h) LASS, (i) RIDGE, (j) NN, (k) SVM, (l) KNN, and (m) STACK.

Another metric that should be discussed is the $R^2$, which is a metric to determine the quality of the model fitting. As seen from Figure 4, $R^2$ is plotted in the y-axis and noise is plotted in the x-axis. By utilizing the same technique of fitting a slope and intercepting onto the results, one can observe the highest $R^2$ estimated at 0% with the intercept and how much the fit is decreased with the slope. A remark is that the intercept (which $R^2$ has a maximum value of 1) is more than 1 due to being an estimate by fitting a line. For linear models such as ELAS, LASS, and RIDGE, there is not a significant decrease of $R^2$ ($R^2$ slope of -0.00, -0.01, -0.10), but this is due to the model being inappropriate for the type of data ($R^2$ intercept of -0.01, -0.01, 0.27). This is the reason why other metrics must be explored coupled with MAE, as the model may underperform from not being a good fit for the data (i.e., non-complex, cannot capture data pattern) compared to other models. Moving on to "tree based" models, all of the models except for ADA (intercept of 0.49) predict the test data extremely well with LGBM, XGB, GB, RF, and DT having an $R^2$ of 0.86, 1.68, 0.97, 0.81, 0.90 respectively. But these models decrease in $R^2$ extremely fast when noise is present in the data (slope of -1.01, -4.12, -1.86, -0.92, -3.90 respectively) with the exception of ADA (slope of -0.52). On to the miscellaneous category, NN, SVM, and KNN have a slope of -0.24, -0.73, and -1.00 and an intercept of 0.35, 0.59, and 0.70 respectively. Again, NN may be underfit as the default parameters may not allow the NN to be trained properly. These models perform moderately in both fitting the data and noise handling. The STACK model is the best of all worlds, as it takes advantage of the high accuracy (intercept of 0.86) of tree-based models and the noise handling (slope of -0.83) of linear and miscellaneous groups, resulting in a good fit and robust model.

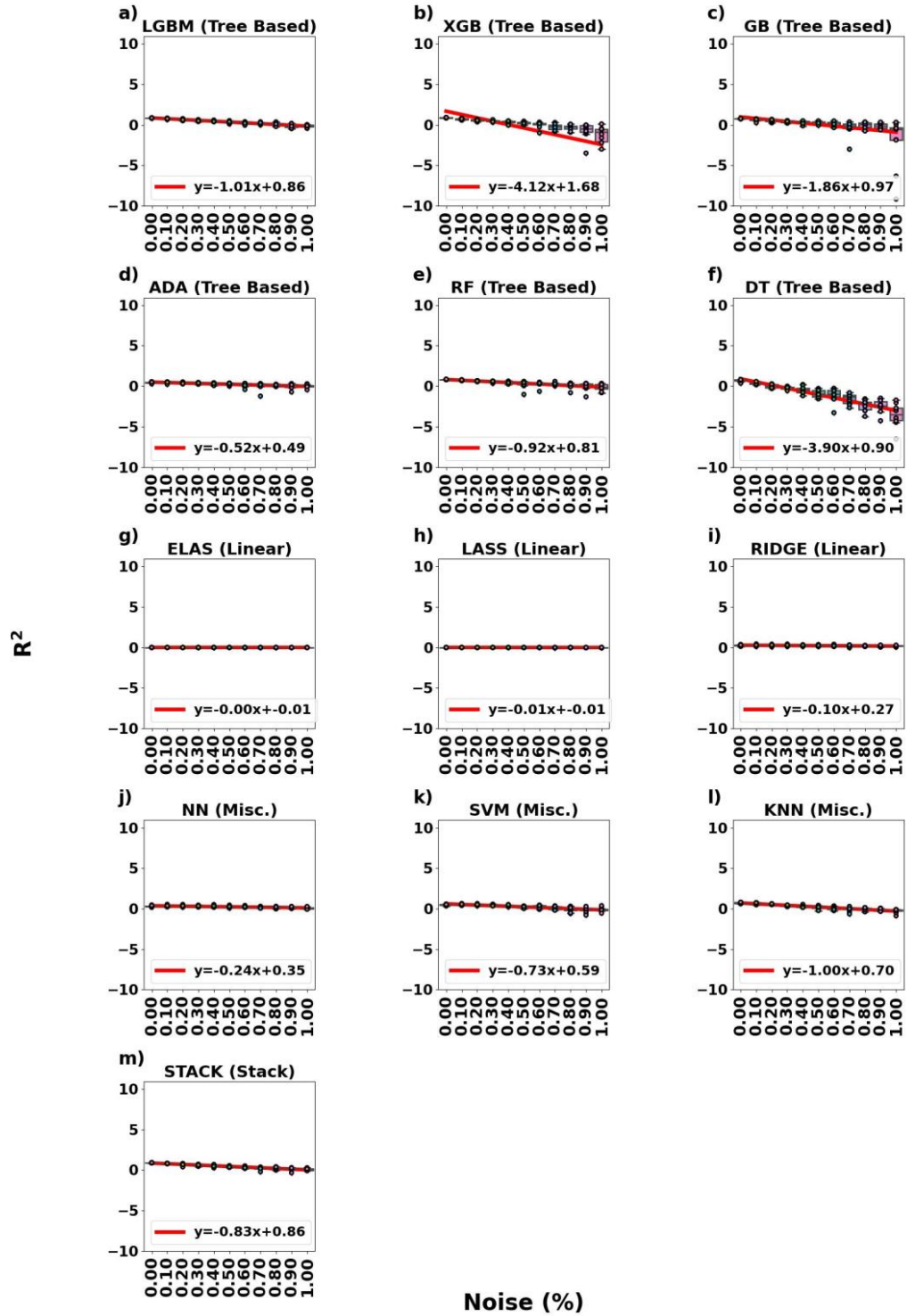

Figure 4: The decrease in $R^2$ of each model when features and target are perturbed for (a) LGBM, (b) XGB, (c) GB, (d) ADA, (e) RF, (f) DT, (g) ELAS, (h) LASS, (i) RIDGE, (j) NN, (k) SVM, (l) KNN, and (m) STACK.

To summarize the results for this section, linear models are robust to noise and provide low accuracy. However, this is due to the improper model selection for this type of data. Tree-based models can achieve high accuracy but are prone to error from noise except for ADA, which has less accuracy but is robust to noise. Miscellaneous models such as SVM, KNN, and NN, are moderately robust to noise and provide moderate accuracy. Even with default parameters, the STACK model performs moderately well in both accuracy and noise handling, which is beneficial to people who just started integrating their electrochemical data with machine learning.

Visualization of feature importance is an important part in the analysis of machine learning results. It can show the region of interest to achieve the desired amount of a prediction. In this case, the objective is to find the region in which the capacitance is the highest, in which a contour plot is one of many techniques for solving this task. To see distinct results, the three models were chosen as representative models to observe how models that are robust to noise affect the contour plot compared to unrobust models. These models are DT (unrobust to noise, moderately accurate), RIDGE (robust to noise, inaccurate), and STACK (moderately robust to noise, highly accurate). Figure 5 is an example of how noise can affect the contour plot between %O and %N for DT. As observed in Figure 5a, the true contour plot of the test set is plotted. The region of high capacitance can be seen around 2% doping of nitrogen and 10% doping of oxygen. This is due to the functionalities present that increase capacitive properties when doped with nitrogen[36-38] (pyridine, pyrrolic, and graphitic forms) and oxygen[39-41] (quinone, phenol). But, not only does doping individually contribute to capacitive properties, they also have synergistic effects when doped together[42, 43]. For example, a study by Barua A. and Paul A. stated that the proton-trapping mechanism of pyridine prevents the decomposition of quinone, which leads to an increase in capacitive properties[43]. This may explain why nitrogen and oxygen co-doping is more beneficial than single doping. Continuing from the model analysis, when the DT model is trained on data with 0% perturbed noise, the prediction contour plot closely resembles the true contour plot as seen in Figure 5b. This suggests that without noise or failure in data mining, the DT model is highly accurate for this dataset. However, when noise is raised to 50%, the contour plot quickly changes to values which do not appear in the original contour plot (i.e., more than 600 F g$^{-1}$), which can be seen in Figure 5c. When 100% noise is present, the contour plot wildly oscillates to extreme ranges (i.e., more than 800 F g$^{-1}$) in Figure 5d. This can confirm how models overfit to noisy data if they are not noise-tolerant. These characteristics are common when using the tree-based techniques (including LGBM, XGB, GB, ADA, and RF). Thus, it is concluded that tree-based models are highly sensitive to noise and can be overfit on the extreme parts of the noisy data. This highlights the importance of finding models that are both highly accurate and robust to noise as the amount of noise is unknown or hard to distinguish in real world datasets.

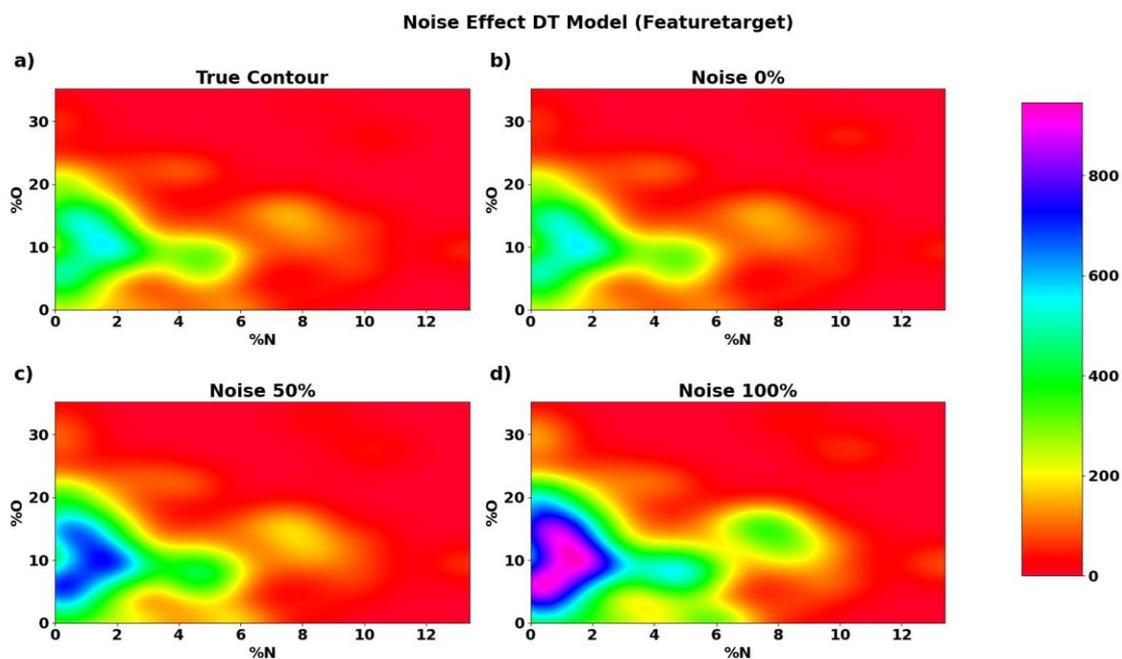

Figure 5: The effect of noise on the contour plot of the test dataset prediction of a DT model where (a) is the contour plot of the true values of capacitance, (b) is the contour plot of the prediction of capacitance when trained on 0% noise, (c) is the contour plot of the prediction of capacitance when trained on 50% noise, (d) is the contour plot of the prediction of capacitance when trained on 100% noise.

In contrast to tree-based models, the linear family shows great noise handling. Models such as RIDGE (see Figure 6) can establish a machine learning model that the prediction is not heavily impacted by noise (false data). As can be seen in Figure 6a and Figure 6b, the prediction contour with 0% noise, and the true value contour plot do not closely resemble each other unlike DT. This is an example of a model having high intercept, which results in inaccuracies in prediction that affects feature importance evaluation. When noise is increased to 50% and 100% as seen in Figure 6c and Figure 6d respectively, the prediction still does not exceed the true contour range, with an exception at 100% noise having slight overestimation. This indicates the robustness of the linear models even at 100% data perturbation. However, it should be noted that the base error from the model prediction is relatively high when compared to tree-based model.

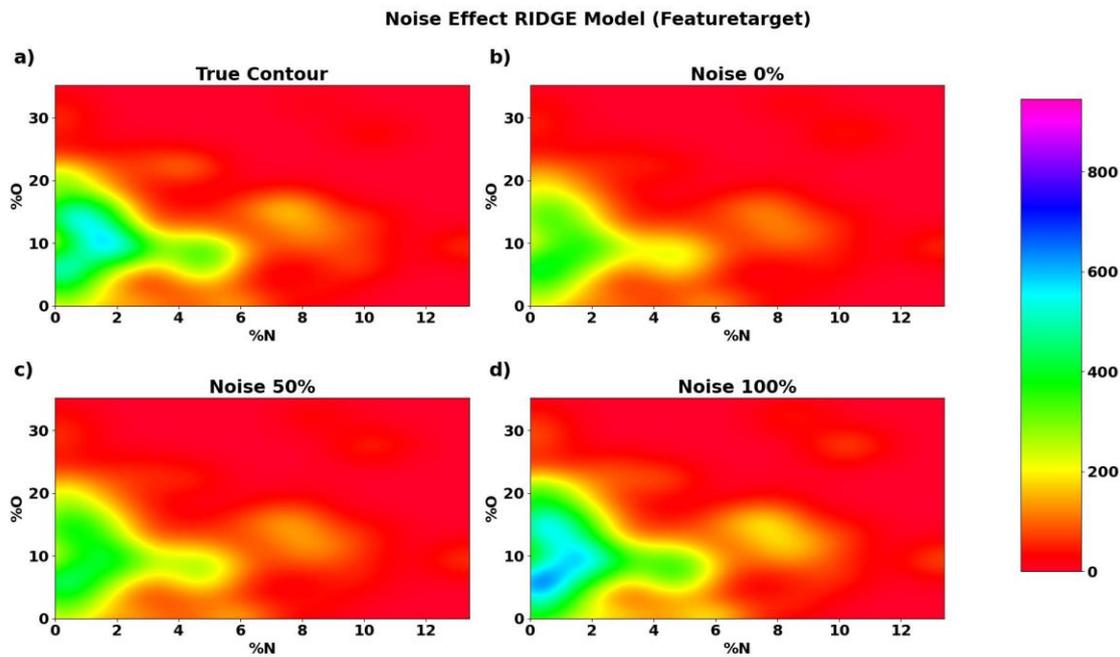

Figure 6: The effect of noise on the contour plot of the test dataset prediction of a RIDGE model where (a) is the contour plot of the true values of capacitance, (b) is the contour plot of the prediction of capacitance when trained on 0% noise, (c) is the contour plot of the prediction of capacitance when trained on 50% noise, (d) is the contour plot of the prediction of capacitance when trained on 100% noise.

From the perspective of the "STACK" model[22], we have presented that the STACK exhibits both high accuracy and robustness. This is due to the combination of model advantages between linear-based and tree-based model, giving the most accurate and robust results. It is evident that the prediction of the stack model with 0% noise shows a closer representation of the true contour than RIDGE but not as close as DT as seen in Figure 7a and Figure 7b. Once there is more false data introduced in, as shown in Figure 7c (50 % error), the "STACK" model can keep the prediction results that stick to the true value. Again, even with 100% noise (see Figure 7d), the contour plot is still close to the true contour and the prediction values do not wildly oscillate. This means that even if there is some error in the as-corrected data due to human errors, error in text mining, or false ("fake") experiments, it will not impact the physical interpretation using "STACK" model unlike standalone tree-based and linear-based models.

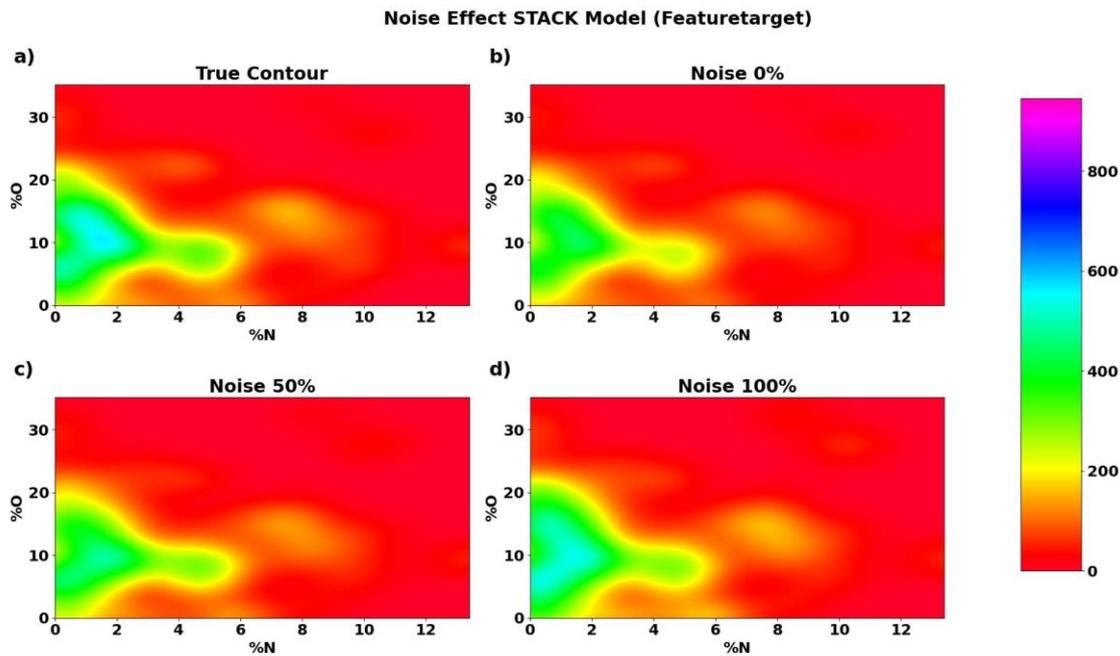

Figure 7: The effect of noise on the contour plot of the test dataset prediction of a STACK model where (a) is the contour plot of the true values of capacitance, (b) is the contour plot of the prediction of capacitance when trained on 0% noise, (c) is the contour plot of the prediction of capacitance when trained on 50% noise, (d) is the contour plot of the prediction of capacitance when trained on 100% noise.

NN should be mentioned as it is a popular model used in this field. However, improperly used NN can result in a model that is either underfitted or overfitted to noisy data. This can be seen in Figure S10 and Figure S11. When the NN model is untuned, a noise-tolerant but underfit model is achieved. In Figure S10b, the model is not yet complex or trained enough to predict the true contour. When noise is increased to 50% and 100% respectively in Figure S10c and Figure S10d, the results still do not oscillate wildly and the range of prediction remains in the true contour region. But over-tuning of the NN model, as seen in Figure S11, can result in a noise-sensitive model. Figure S11b shows the prediction at 0% noise, which still remains in range of the true contour plot. But when noise is increased to 50% and 100% as seen in Figure S11c and Figure S11d, the contour plot starts to wildly vary (more than 800 F g$^{-1}$). This heavily impacts feature importance analysis and conclusions drawn from the model when noise is high. For additional model analysis and demonstration of how models fit data, refer to the "Analysis of Model Performance" and "Demonstration of Model Abilities" section in the Supplementary Information.

In addition to contour plots, SHAP analysis is a way to observe feature importance by using game theory, which is typically used in this field[14, 16, 22, 44, 45]. The widely used Kernel SHAP was developed by Scott Lundberg and Su-In Lee to provide a computationally inexpensive and accurate approximation of the Shapley values[46]. The higher the Shapley values, the more that feature contributes to the prediction (which is the capacitance). The first 100 values of the reserved test set are used to evaluate SHAP. From various literatures, it is observed that higher surface area has a positive correlation with capacitance, as reported in previous literatures[47, 48]. This is due to the charge storage mechanism in supercapacitors, which rely on physical adsorption (physisorption) to the surface of the material[49, 50]. Therefore, increasing the surface area will increase the physisorption, resulting in higher capacitance contributed by Helmholtz capacitance. Note that the Helmholtz capacitance is calculated by multiplying the vacuum permittivity ($\varepsilon_0$ = 8.854 x 10$^{-12}$ Fm$^{-1}$), relative permittivity of electrolyte ($\varepsilon_r$), and the exposed surface area (S) then divided by the charge separation length (d).

$$C_H = \frac{\varepsilon_0 \varepsilon_r S}{d} \qquad \text{(Eqn. 6)}$$

As can be seen from Eqn 6, the Helmholtz capacitance is directly proportional to the exposed surface area ($C_H \propto S$). This indicates a linear relationship, and simple models which fit a line such as linear regression (LIN), ELAS, LASSO, and RIDGE can be used. This behavior can be observed in the SHAP analysis of this study in Figure 8, which shows that at 0% noise, the SHAP results of DT, RIDGE, and the STACK model can replicate linear relationships. But, when the noise is increased to 50%, DT's SHAP seems to wildly oscillate and does not resemble a linear relationship unlike at 0% noise. In contrast, RIDGE and STACK still perform well and exhibit positive linear relationships. At the extreme end of 100% noise, RIDGE can still approximate a positive linear relationship true to the Helmholtz equation while STACK is overwhelmed by the base models that underperform, resulting in a flat SHAP. This suggests that overly complex models like DT and even STACK with multiple models may not be recommended for strictly simple linear relationships assuming large quantities of noisy data.

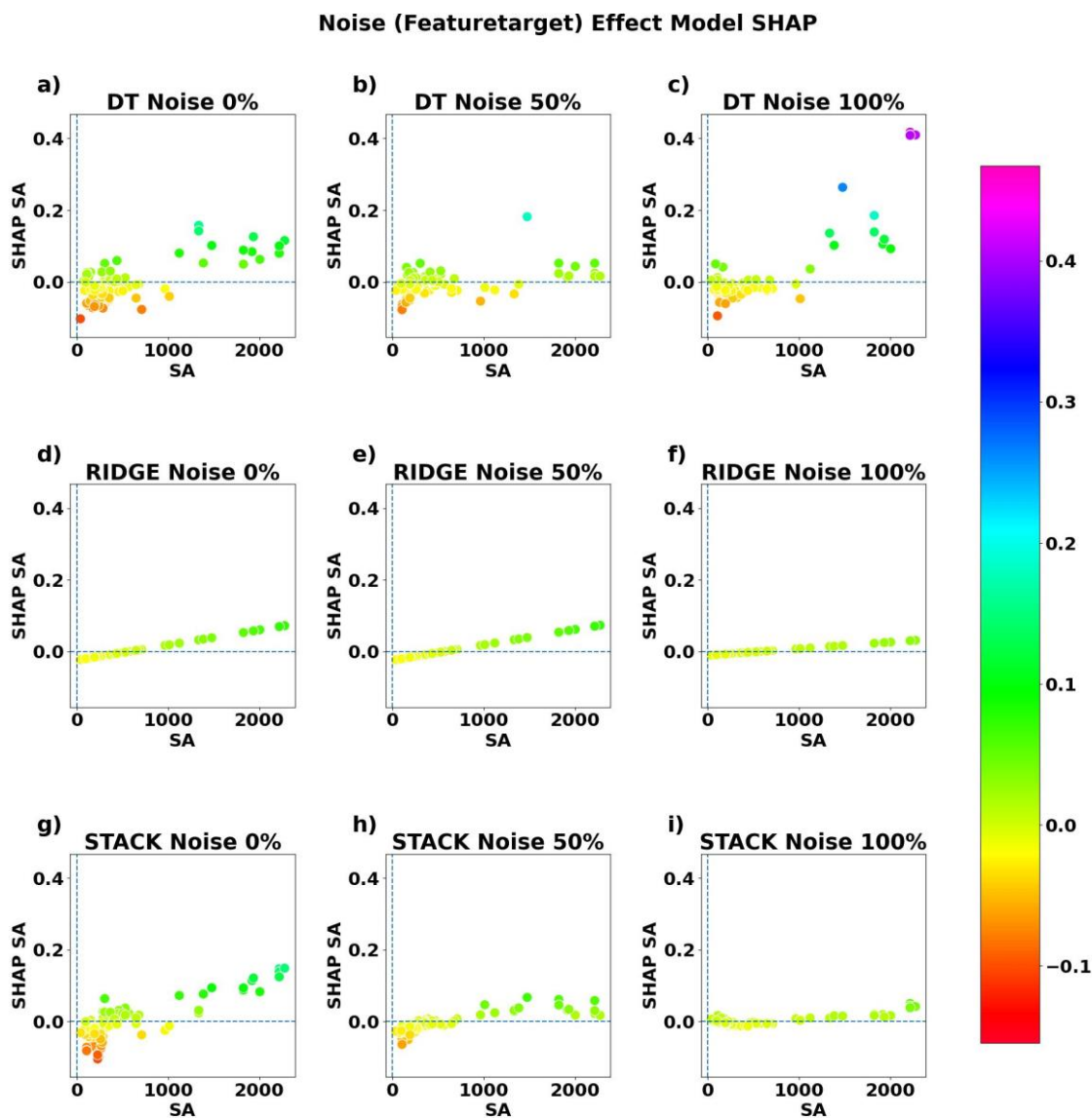

Figure 8: The effect of noise on surface area SHAP of DT where the model is trained on (a) 0% noise, (b) 50% noise, and (c) 100%. The effect of noise on surface area SHAP of RIDGE where the model is trained on (d) 0% noise, (e) 50% noise, and (f) 100%. The effect of noise on surface area SHAP of STACK where the model is trained on (g) 0% noise, (h) 50% noise, and (i) 100%.

For more complex interactions, for example, current density and capacitance, simple linear relationships cannot fully capture the pattern in the data, as the graph seems to resemble a log function with a fraction base[51, 52]. To the best of our knowledge, there is no fundamental equation relating current density and capacitance, therefore, more complex models such as polynomial regression (POLY), SVM, DT and its variants (e.g., XGB, ADA, RF) are used instead to capture the relationship and provide insight into how current density changes capacitance. The results of SHAP of current density can be observed in Figure 9. The results suggest that at 0% noise DT and STACK can appropriately replicate the current density graphs, while RIDGE struggles to replicate since it is a simple and non-complex linear model, but shows a decreasing trend from increased current density, which is still a valuable insight into the mechanisms of current density and capacitance. When the noise is increased to 50%, DT and STACK still perform well and can show a curved trend while RIDGE still shows decreasing trends. On the extreme end of 100% noise, DT's performance seems to deteriorate while STACK continues to maintain the curved trend. RIDGE's trend still continues down similar to 0% noise. The results can be validated back to the experimental data including plots of capacitance vs. surface area and capacitance vs. current density.

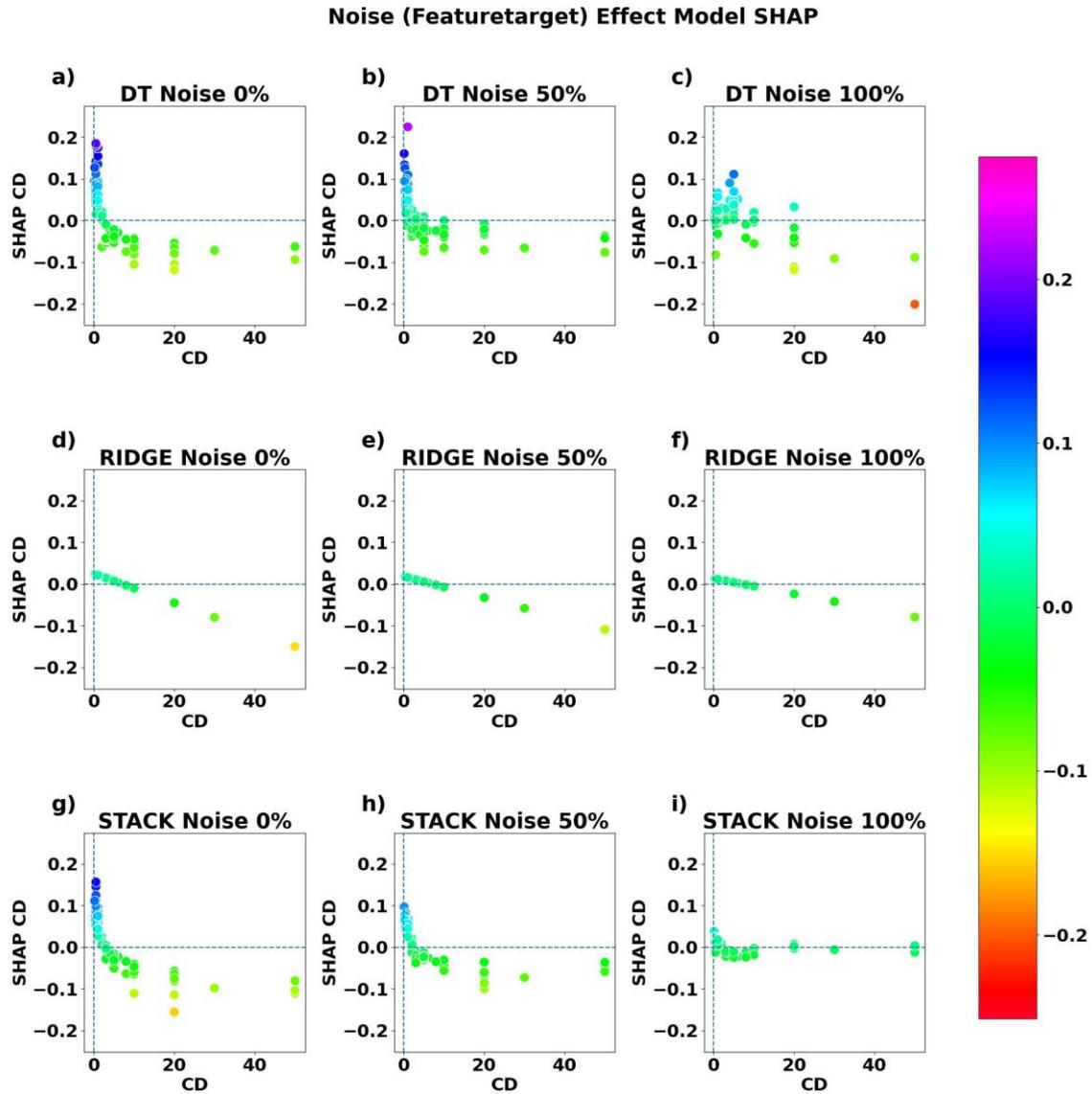

Figure 9: The effect of noise on current density SHAP of DT where the model is trained on (a) 0% noise, (b) 50% noise, and (c) 100%. The effect of noise on current density SHAP of RIDGE where the model is trained on (d) 0% noise, (e) 50% noise, and (f) 100%. The effect of noise on current density SHAP of STACK where the model is trained on (g) 0% noise, (h) 50% noise, and (i) 100%.

However, the difficulties may occur when there is no former trend to compare or studies that describe the interaction. This is the case for oxygen doping, as no ground truth has been established for how much oxygen should be doped into graphene to obtain the highest capacitance. By revisiting the scatter plot between percent of doped oxygen and capacitance in Figure S1, no clear trends can be seen, and the data suggests heavy amounts of noise in 2D space. But, when SHAP is applied, as shown in Figure 10, some peaks can be seen where optimal amounts of oxygen doping is favorable. At 0% noise, STACK shows a trend with a peak at roughly 15% doped oxygen. This means that to obtain the highest capacitance, oxygen should roughly be doped in the 15% range. This suggests that under-doping or over-doping oxygen may degrade the capacitance of the supercapacitor due to there not being enough active oxygen species such as quinone, which is an excellent redox contributor that can transform into hydroquinone[53], and there being more acidic surface functionalization species such as carboxyl groups, which can reduce capacitive properties due to the inhibition of ion movement when charging and discharging[54]. Unfortunately, the selection of specific functionalization is hard to control, which is why optimal conditions are present. Continuing, DT seems to have some data points that are skipped higher at 15% oxygen, but the SHAP line seems to be flat. RIDGE seems to provide a negative trend, but the relationship may be too complex for linear models. When increasing the noise to 50%, DT shows a better trend with peaks at 10% to 20% oxygen, RIDGE seems to show a largely negative trend, and STACK's result is drowned out. All this can be due to a lucky/unlucky randomization seed. At 100% noise, DT seems to show the most trend, while RIDGE and STACK is a flat trend line. In the Supplementary Information, additional testing of another technique called partial dependence plots can be found in Figure S15 to Figure S17.

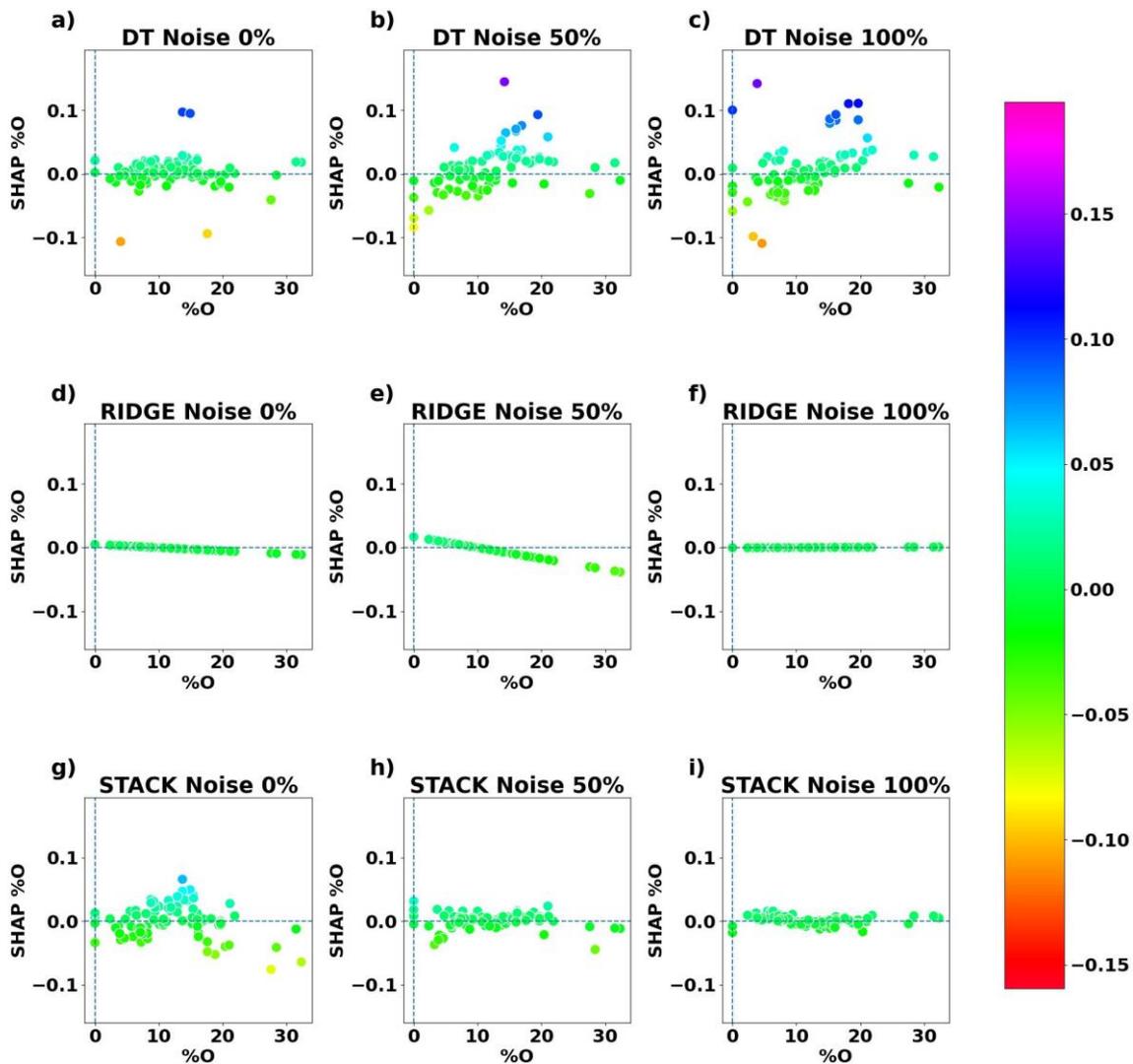

Figure 10: The effect of noise on oxygen SHAP of DT where the model is trained on (a) 0% noise, (b) 50% noise, and (c) 100%. The effect of noise on oxygen SHAP of RIDGE where the model is trained on (d) 0% noise, (e) 50% noise, and (f) 100%. The effect of noise on oxygen SHAP of STACK where the model is trained on (g) 0% noise, (h) 50% noise, and (i) 100%.

Other than noise, falsified or inflated data can cause incorrect conclusions to be drawn and reduce the accuracy of models. Referring to a study done by Fanelli D., it is established that roughly 34% of scientists report suspicious data handling practices[55]. This leads to suspicions as to whether this amount of data, which may be "inflated" slightly, will cause a problem for machine learning in electrochemistry. Therefore, tests were conducted as to what effect inflated data has on DT, RIDGE, and STACK. The workflow for the process is similar to Figure 11, where data is perturbed, but only for the target variable. To get a baseline to compare, the model is trained on normal data (i.e., 0% added noise) and the prediction is obtained. To simulate inflated data, 34% of the training data is randomly sampled and inflated by 7.5% to 15%. This inflated data is then used to train the model and the result is compared, where the real value is plotted in the x-axis and the predicted value is plotted in the y-axis. Observing when DT is trained with inflated data in Figure 11a and with normal data in Figure 11b, it can be seen that the $R^2$ is decreased when the model is trained on inflated data and more predicted datapoints are scatted above the perfect prediction line (93 versus 58), indicating that the inflated data causes a skewness in prediction. Kernel Density Estimation (KDE) can also be used to get a rough estimate of the distribution density, as the density for inflated data is above the perfect prediction line rather than in the middle. The same can be said for RIDGE and STACK, but a key difference is the reduction in $R^2$ when trained on inflated data and normal data. For DT, the reduction is 0.0550, while RIDGE is 0.0479 and STACK is 0.0293. This shows that the STACK model can handle being trained with inflated data compared to DT and RIDGE, as $R^2$ for STACK is least affected. Finally, STACK has a higher $R^2$ (0.7526 $R^2$ for normal data and 0.7233 $R^2$ for inflated data) compared to DT (0.6788 $R^2$ for normal data and 0.6238 $R^2$ for inflated data) and RIDGE (0.6145 $R^2$ for normal data and 0.5666 $R^2$ for inflated data) in both normal and inflated data. All this indicates that not only can STACK handle falsified inflated data, but it can also provide a good fit to graphene supercapacitor data.

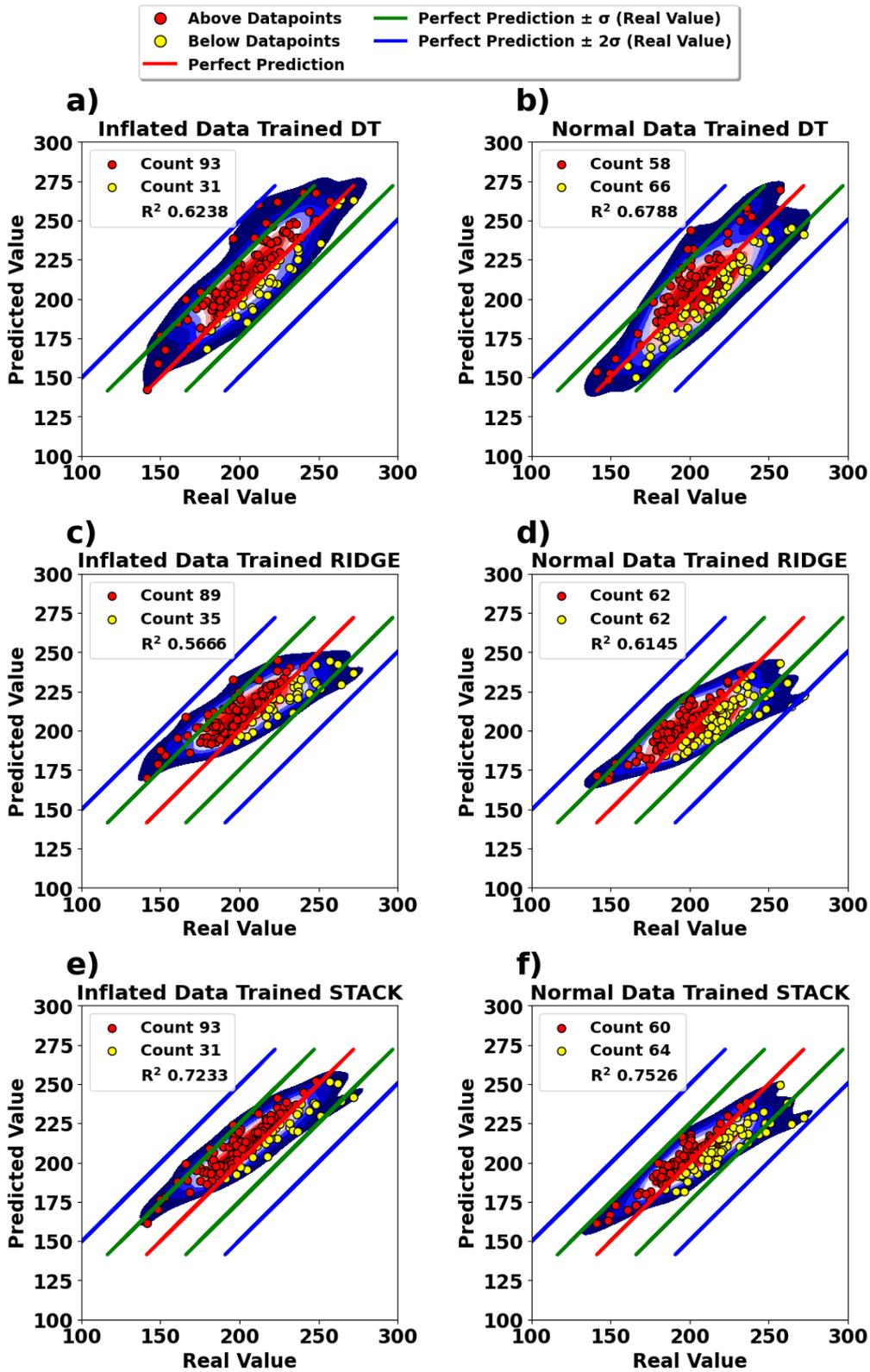

Figure 11: Comparison of models trained on 34% of total data inflated by 5% to 10% for DT on (a) inflated data, (b) normal data. For RIDGE on (c) inflated data, (d) normal data. For STACK on (e) inflated data, (f) normal data

CONCLUSION

False data (also known as "noise", and "error") was introduced into the feature, target and both feature-target variables of datasets from 0% to 100% with a 10% interval to gain an understanding of how models behave when trained with noisy data. A total of 12 models which includes XGB, LGBM, RF, GB, ADA, NN, ELAS, LASS, RIDGE, SVM, KNN, and DT were used. The training included using 10 random split seeds to increase reliability in results. The findings of this study are summarized in a radar plot in Figure S18. The standalone tree-based model is highly susceptible to noise (DT MAE slope of 102.11 F $g^{-1}$) but shows an acceptable ability to achieve accurate predictions (DT MAE intercept of 32.29 F $g^{-1}$). Other tree-based models that utilize bagging and boosting (e.g., XGB, LGBM, RF, GB, ADA) increase the accuracy while also providing tolerance to noise. The most tolerant-to-noise tree-based model is ADA with a MAE slope of 16.84 F $g^{-1}$, but in contrast to other tree-based models, ADA suffers from prediction accuracy with an MAE intercept of 47.70 F $g^{-1}$. These tree-based models also fit well to the data (average $R^2$ of 0.9516). Linear models exhibit the highest ability to handle noise, however, it has a trade-off of prediction accuracy, as the highest errors are present in these models (average MAE intercept of linear models is 60.20 F $g^{-1}$). This is most likely due to their inability to fit to complex data (average $R^2$ of 0.25). Miscellaneous models such as SVM, KNN, and NN, are moderately robust to noise (average MAE slope of 25.956 F $g^{-1}$) and provide moderate accuracy (average MAE intercept of 41.306 F $g^{-1}$) with the exception of NN (9.68 and 49.71 F $g^{-1}$ MAE slope and intercept), which may be underfit from the default parameters. The miscellaneous models also fit moderately well to the data (average intercept of 0.546 $R^2$). The STACK model performs exceptionally in accuracy (24.29 F $g^{-1}$ MAE intercept), moderately well in noise handling (41.38 F $g^{-1}$ MAE slope), and achieves a high fit (0.86 $R^2$ intercept) by combining the advantage of all models. This error is relatively close to the experimental data when performing more replicates. Feature importance analysis techniques (i.e., contour plot, SHAP, PDP) performed on models trained with noisy data may not be accurate as they can overfit or underfit to the noisy data. This study aims to show that even with untuned models, selection of the best model to handle noisy data is critical to an accurate prediction and reliable feature importance analysis. A way to minimize risk of error is to stack available models and to compare their results to standalone ones before tuning the model. Aleatoric error must be considered apart from epistemic error (hyperparameter tuning) when performing model selection. This work should improve the understanding of using machine learning in not just for energy storage but chemistry area.

AUTHOR INFORMATION

*Corresponding Author

P. Iamprasertkun* (pawin@siit.tu.ac.th) Tel: +66-2-986-9009 ext. 2306

ORCID: https://orcid.org/0000-0001-8950-3330

Krittapong Deshsorn (work.krittapongd@gmail.com)

ORCID: https://orcid.org/0009-0008-5251-8625

Author Contributions

K.D.: methodology, software, data curation, formal analysis, visualization, and writing-original draft; L.L.: Resources, Supervision; P.I.: conceptualization, validation, formal analysis, writing─original draft, review, and editing, supervision, project administration, and funding acquisition. All authors have given approval to the final version of the manuscript.

Funding Sources

SIIT Young Researcher Grant (under contract No. SIIT 2023-YRG-PI01)

ACKNOWLEDGMENT

This work is supported by the SIIT Young Researcher Grant (under contract No. SIIT 2023-YRG-PI01). Authors acknowledge the facilities from Sirindhorn International Institute of Technology, Thammasat University. Krittapong Deshsorn acknowledges the ETS scholarship awarded by Sirindhorn International Institute of Technology, Thammasat University.

ABBREVIATIONS

SHAP, Shapley Additive Explanations; SA, surface area; DG, defect ratio (ID/IG); %N, percentage of doped nitrogen; %O, percentage of doped oxygen; %S, percentage of doped sulphur; CD, current density; CONC, electrolyte concentration; CAP, capacitance; LGBM, Light gradient boosting machine; XGB, Extreme gradient boost; NN, Neural network; ELAS, Elastic Net Regression; LASS, Lasso regression; RIDGE, Ridge regression; RF, Random forest; SVM, Support vector machine; KNN, K-nearest neighbors; GB, Gradient boost; ADA, Adaboost; DT, Decision tree; CSV, Comma separated value; STACK, Stacking model; PDP, Partial dependence plot